\documentclass[10pt,twocolumn,letterpaper]{article}
\PassOptionsToPackage{table}{xcolor}
\usepackage[pagenumbers]{cvpr}

\usepackage{xcolor}
\usepackage{makecell}
\definecolor{cvprblue}{rgb}{0.21,0.49,0.74}
\definecolor{bestred}{RGB}{146,183,219}
\definecolor{secondorange}{RGB}{208,225,242}
\newcommand{\best}[1]{\cellcolor{bestred}#1}

\usepackage[pagebackref,breaklinks,colorlinks,allcolors=cvprblue]{hyperref}
\def\confName{CVPR}
\def\confYear{2027}
\title{TokenMatte: Unlocking T2V Diffusion Priors for Pixel-Native Video Matting}
\author{
Zanyi Wang$^{1}$ \qquad
Dengyang Jiang$^{2}$ \qquad
Fengfei Yu$^{1}$ \qquad
Shilong Liu$^{3}$\\[6pt]
$^{1}$UC San Diego \qquad
$^{2}$HKUST \qquad
$^{3}$Princeton University
}
\begin{document}
\maketitle

\begin{abstract}
High-resolution video matting requires accurate recovery of continuous alpha values, including fine hair, soft boundaries, and motion blur, yet dense alpha supervision remains scarce. Pretrained text-to-video diffusion transformers (DiTs) offer rich spatio-temporal priors, whereas existing generative video matting approaches typically route alpha predictions through RGB-trained VAE latents, which can attenuate fine details and introduce additional decoding overhead. We present TokenMatte, which adapts a pretrained video DiT into a single-pass, auxiliary-free RGB-to-alpha predictor with a pixel-native output interface. At its core is a single shared linear readout that maps each spatio-temporal token directly to its corresponding $(t \times p \times p)$ block of alpha pixels, predicting spatial details and temporal variations within the same local video block. Tiling these blocks produces the full-resolution alpha sequence without a convolutional decoder or learned upsampling. The predicted alpha is supervised directly in pixel space, while the ground-truth alpha never enters the VAE, eliminating target-side alpha encoding and output-side VAE decoding. Leveraging the pretrained video prior, TokenMatte is trained using only existing public matting datasets, without additional data curation or auxiliary inputs at inference. Compared with a matched latent-output baseline, our pixel-native interface improves matting accuracy while achieving a $1.51\times$ inference speedup. TokenMatte establishes a new auxiliary-free state of the art on VideoMatte, YouTubeMatte, and V-HIM60.
\end{abstract}

\begin{figure*}
    \centering
    \includegraphics[width=1\linewidth]{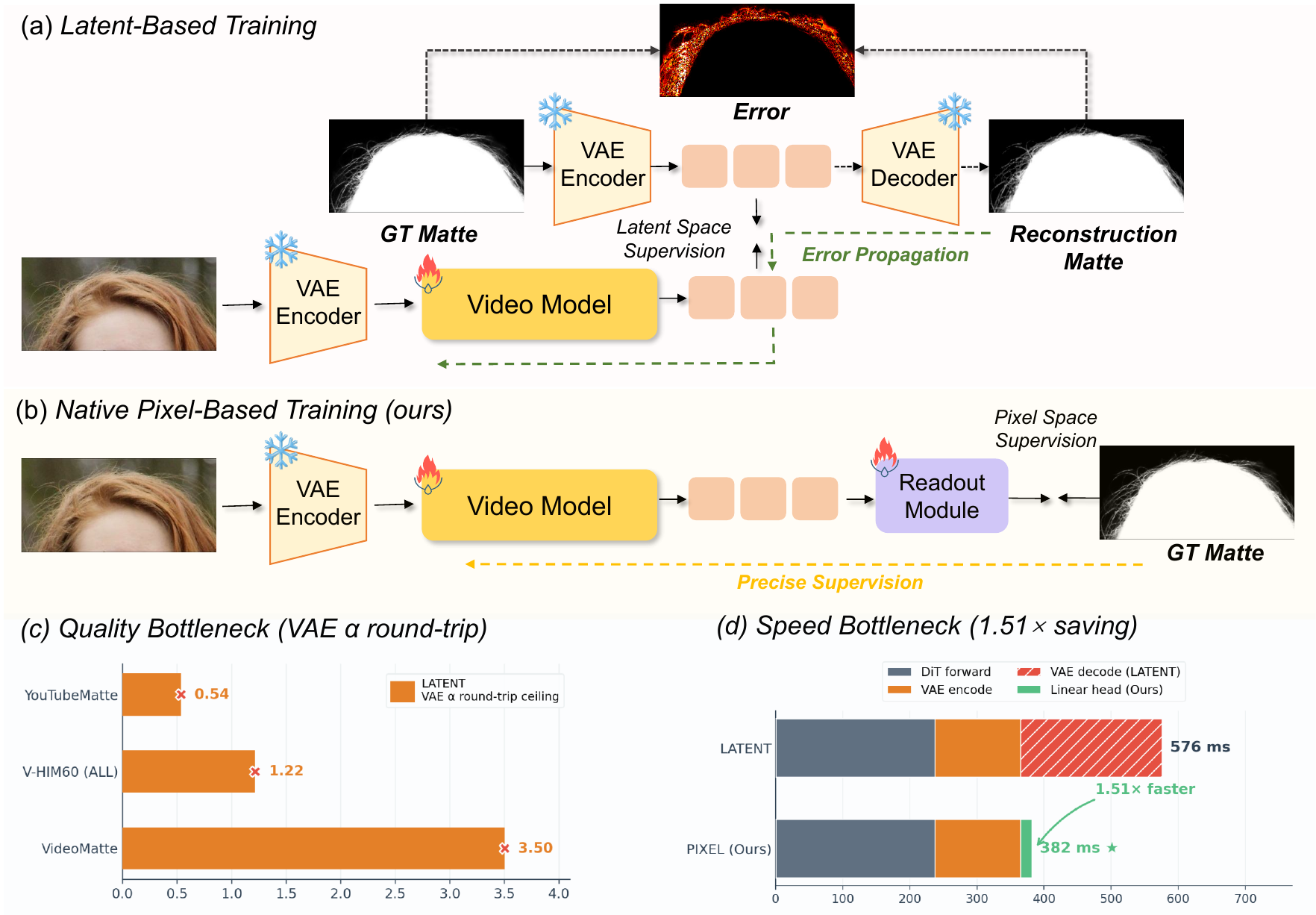}
  \captionof{figure}{\textbf{Latent-based vs.\ our native pixel-based training.}
  \textbf{(a)} Latent supervision encodes the GT matte through a frozen
  VAE and supervises in latent space; the alpha target must survive a
  VAE round-trip, and the pixel objective only reaches the model through
  the decoder. \textbf{(b)} We never encode the alpha target: a readout
  module unfolds each token directly into pixels, supervised in pixel
  space. \textbf{(c)} Quality bottleneck---the GT matte incurs reconstruction error after a VAE round-trip, revealing an empirical ceiling for alpha reconstruction, especially on detail-heavy data (VideoMatte $3.50$, V-HIM60 $1.22$, YouTubeMatte $0.54$, MAD). \textbf{(d)} Speed bottleneck---removing the VAE-decode stage cuts per-clip latency from $576$\,ms to $382$\,ms,
  a $1.51\times$ saving at equal depth. }
  \label{fig:latent_vs_pixel}
\end{figure*}
\section{Introduction}
\label{sec:intro}

Video matting is a core component of modern video editing, compositing, background replacement, and visual effects. In the most practical setting, the user provides only an RGB video and expects a temporally coherent alpha sequence at the original resolution, without trimaps, masks, point prompts, or first-frame annotations~\cite{chuang2002video,apostoloff2004bayesian,sengupta2020background,lin2022robust}. This auxiliary-free setting minimizes user interaction and simplifies deployment, but it is also one of the most challenging operating points for matting: the model must recover fine boundary transitions, hair structures, motion blur, transparent regions, and foreground semantics directly from the video without any additional cues~\cite{ke2022modnet,lin2022robust}.

A central bottleneck is supervision. Unlike segmentation masks, alpha mattes are continuous pixel-space quantities that encode fractional opacity. A mask can identify foreground regions, but it does not directly supervise hair-level structures, semi-transparent boundaries, motion-blurred transitions, or soft foreground-background mixtures~\cite{cai2019disentangled}. Accurate alpha labels are therefore expensive to annotate at scale, especially for real high-resolution videos with long temporal variation~\cite{xu2017deep,yang2026matanyone}. As a result, much of the publicly available video matting supervision is green-screen captured, composited, synthetic, pseudo-labeled, short, or limited in scale~\cite{lin2021real,wang2021video,huynh2024maggie}. This data bottleneck is particularly severe for auxiliary-free matting, where no external mask or reference annotation is available at inference time.

Recent video matting research has actively sought to overcome this bottleneck~\cite{yang2025matanyone,yang2026matanyone,lim2026videomama,ge2025generative}. One line of work improves matting by scaling or curating task-specific data, including large real-video or pseudo-labeled video matting datasets, learned quality evaluators, and mask-to-matte pipelines~\cite{yang2025matanyone,yang2026matanyone,lim2026videomama,huynh2024maggie}. These methods have substantially advanced the field, but they also shift progress toward increasingly large self-curated data resources and additional guidance signals. Another line of work exploits generative video priors, showing that pretrained video generation models can help bridge the gap between limited alpha supervision and real-world appearance variation~\cite{ge2025generative,wang2025deforming}. However, supervising in this lossy latent space limits matting performance because the VAE round-trip destroys fine-grained details (e.g., hair) and propagates reconstruction errors during training; even an oracle round-trip of the ground-truth matte through the VAE already loses fine detail, and this ceiling is largest on the most detail-heavy benchmarks (Fig.~\ref{fig:latent_vs_pixel}(a),(c)).

In this work, we present TokenMatte, which takes a different path to bypass the data bottleneck. Instead of scaling task-specific data or relying on auxiliary prompts, we directly adapt a well-pretrained video DiT into an RGB-to-alpha predictor. Since a large-scale video DiT already organizes video inputs into rich spatio-temporal tokens that encode appearance, motion, and temporal continuity~\cite{chen2025video,wan2025wan}, extracting the alpha matte becomes a ``readout" task of an inherited representation rather than a heavy synthesis process. Publicly available alpha data, without any self-curated resources, is fully sufficient to teach this translation.

To exploit this prior, existing methods typically follow standard diffusion models by applying supervision in the latent space~\cite{ge2025generative,lim2026videomama}, specifically,  compressing the ground-truth alpha through a video VAE and supervising the model in latent space. However, as illustrated in Fig.~\ref{fig:latent_vs_pixel}(a) and (c), even a simple VAE ``round-trip" (compressing the ground-truth matte into the latent space and decoding it back) incurs significant reconstruction error and information loss, revealing a strong empirical reconstruction ceiling for alpha mattes. Supervising the model within this lossy latent space inevitably propagates these VAE-induced errors back to the video backbone during training (Fig.~\ref{fig:latent_vs_pixel} (a)), compounding the optimization difficulty and preventing the network from capturing true pixel-level boundaries.  We show that the target interfaces of existing generative methods are unnecessarily indirect, and circumvent this limitation by directly supervising the model in pixel space (Fig.~\ref{fig:latent_vs_pixel}(b)). Specifically, we leverage the DiT's fixed patch-aligned token
grid, where each spatio-temporal token corresponds to a fixed local region, and employ a lightweight readout module to map tokens directly to alpha pixels.

This native pixel-space training enables pristine supervision, preserves delicate boundaries, and drastically improves efficiency. By completely bypassing the expensive VAE-decoding stage during inference, we eliminate a major computational bottleneck, reducing per-clip latency by 
1.51 $\times$  (Fig.~\ref{fig:latent_vs_pixel}(d)). Ultimately, this design clarifies the roles of the prior and the data: pretraining provides the spatio-temporal foundation, while public alpha data merely calibrates the token-to-alpha translation, unlocking the DiT as an efficient, RGB-only, alpha-native predictor.

In summary, our main contributions are as follows:

\begin{itemize}
\item We present TokenMatte, which adapts a pretrained video DiT into an
auxiliary-free matting model, while retaining a lightweight
token-to-pixel output interface.
\item We show this translation needs only a single per-token linear readout that decodes each spatio-temporal token into its $(t\times p\times p)$ block of alpha pixels---no alpha encoding, latent denoising, or VAE decoding.
\item Our model sets a new auxiliary-free state of the art on VideoMatte, YouTubeMatte, and V-HIM60, and runs 1.51× faster than the latent interface baseline.
\end{itemize}

\section{Related Work}
\label{sec:related}

\subsection{Auxiliary-free video matting}
Video matting predicts, for every frame of an RGB video, an alpha matte that is accurate within each frame and stable over time. It is closely related to segmentation but harder: rather than a discrete label per pixel, matting estimates a continuous opacity that captures the soft transitions of hair, motion blur, and semi-transparent boundaries~\cite{xu2017deep,hou2019context}, and unlike its image counterpart~\cite{lu2019indices,li2020natural,park2022matteformer}, it must also stay temporally coherent, since individually plausible frames can still flicker as a sequence~\cite{li2021deep,li2021privacy,zhang2021attention}. To reduce this ill-posedness, most methods take an auxiliary input: a per-frame trimap, costly to annotate, or a single first-frame mask~\cite{xu2017deep}. MaGGIe mattes human instances from binary mask tracks, MatAnyone propagates a target-assigned alpha through memory, MatAnyone-2 adds a learned quality evaluator and a large self-curated training set, and VideoMaMa conditions a generative prior on first-frame masks from SAM2; all reintroduce annotation or tracking at inference~\cite{huynh2024maggie,yang2025matanyone,lim2026videomama,kirillov2023segment,yu2021mask}.

We instead focus on auxiliary-free matting, where the model sees only the RGB input and predicts alpha with no trimap, mask, prompt, or annotation. This is the most convenient and deployable setting, and a long line of image and video methods pursue it directly~\cite{ke2022modnet,lin2022robust,li2021privacy,li2021deep,ge2025generative}. Its central difficulty is data: dense alpha labels are far more costly than segmentation masks, and public matting supervision is largely composited, synthetic, or limited in scale~\cite{lin2021real,huynh2024maggie,yang2026matanyone}, which is precisely the gap a pretrained generative prior can help fill.

\subsection{Generative priors for video matting}
Large pretrained generative models have emerged as powerful visual priors, capturing appearance, structure, and, for video, motion and temporal continuity from large-scale synthesis~\cite{blattmann2023stable,yang2025cogvideox,zheng2024open}. Early efforts built on latent-diffusion U-Nets~\cite{rombach2022high}, which reach high resolution only by repeatedly downsampling into convolutional feature maps; modern diffusion transformers instead scale a video into a flat lattice of spatio-temporal tokens, each bound to a fixed space-time patch~\cite{blattmann2023stable,wan2025wan}. Riding this prior, dense prediction has been recast as generation: latent-diffusion priors have been adapted to depth, normals, image matting, and segmentation masks~\cite{fu2024geowizard,he2025lotus,ke2024repurposing,xu2025matters,yang2026genmask}, and for video matting GVM~\cite{ge2025generative} casts matting as flow-matching in latent space while VideoMaMa~\cite{lim2026videomama} synthesizes alpha-matte latents from a video diffusion backbone. Across this line, however, the alpha target is reached only through the backbone's VAE: it is encoded into an RGB-trained latent and supervised there, so an opacity field scored in pixel space must survive an 8x-compressed space optimized for natural images, eroding the sub-pixel hair and semi-transparent boundaries matting must recover while adding a decode pass at high resolution~\cite{ge2025generative,lim2026videomama,li2026back,yao2025reconstruction}. Building on the ViT patch paradigm, pixel-space generators like
JiT~\cite{li2026back} share DiT's~\cite{peebles2023scalable} per-patch
token view but drop the VAE to read raw pixels directly, an insight our
readout extends to auxiliary-free, high-resolution video matting by
mapping each video DiT token to its corresponding block of alpha pixels.
\section{Method}
\label{sec:method}

\begin{figure*}[t]
  \centering
  \includegraphics[width=\linewidth]{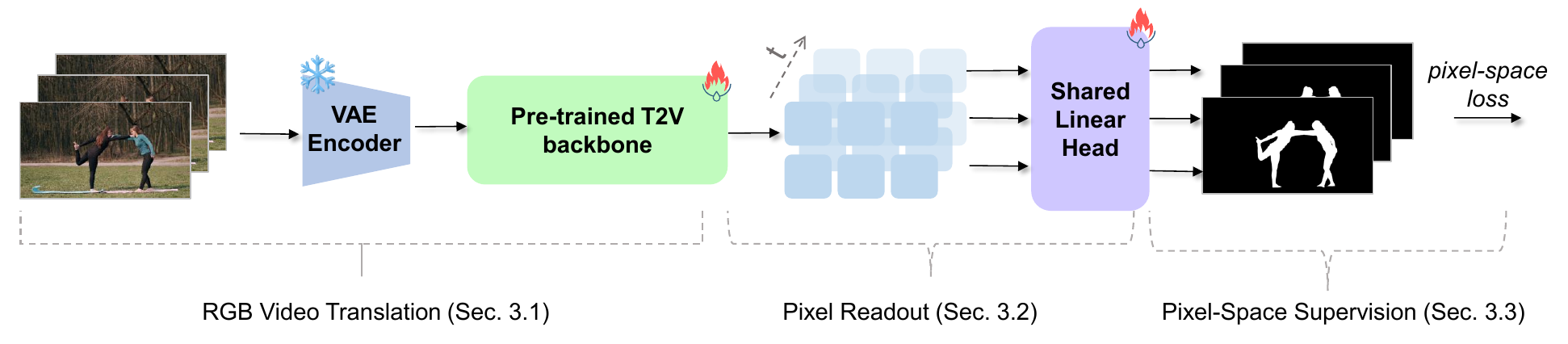}
  \caption{\textbf{Method Overview.} From an RGB video, a frozen VAE encoder
  produces latent tokens and a pretrained T2V backbone runs one
  deterministic pass ($\tau\!=\!\tau_0$, $\epsilon\!=\!0$) to yield a
  spatio-temporal token field (Sec.~\ref{sec:method:translation}). A
  shared linear head unfolds each token into its local block of alpha
  pixels (Sec.~\ref{sec:method:readout}), supervised in pixel space with
  no alpha encoding or VAE decoding (Sec.~\ref{sec:method:novae}).}
  \label{fig:overview}
\end{figure*}

Recent generative video matting reaches the alpha target through the backbone's VAE: alpha is encoded into a latent, supervised there, and decoded back to pixels~\cite{ge2025generative,lim2026videomama}. This inherits two costs from RGB synthesis. The VAE is optimized to reconstruct natural RGB, not the continuous transmittance of an alpha matte, so encoding alpha through it compresses exactly the sub-pixel hair and soft boundaries matting must recover, capping accuracy before learning begins; and decoding back to pixels adds a pass whose cost is most severe at the high resolutions matting demands. We avoid both by never sending alpha through the VAE: we unlock the RGB prior of a pretrained video DiT for video matting and read the alpha field directly off its tokens, in one forward pass. Following Fig.~\ref{fig:overview} from left to right, we utilize the prior (Sec.~\ref{sec:method:translation}), read alpha off the token grid (Sec.~\ref{sec:method:readout}), and supervise it in pixel space (Sec.~\ref{sec:method:novae}). We seek the alpha sequence $\boldsymbol{\alpha} \in \mathbb{R}^{T \times 1 \times H \times W}$ from an RGB video $\mathbf{x} \in \mathbb{R}^{T \times 3 \times H \times W}$, with no trimap, mask, prompt, or first-frame annotation.

\subsection{RGB Video Translation}
\label{sec:method:translation}

\textbf{Reusing the pretrained RGB interface.} The RGB video enters through the backbone's frozen VAE encoder $\mathcal{E}$, preserving the pretrained RGB input interface. The video DiT is then fine-tuned for RGB-to-alpha prediction, while the alpha target never passes through the VAE (Sec.~3.3).

\paragraph{From generation to a single-pass translation.}
A video DiT $F_\theta$ is pretrained for conditional generation, mapping a noised latent at timestep $\tau$ to a velocity or noise estimate and sampling by iterative denoising. For a deterministic dense target this stochastic interface is unnecessary~\cite{xu2025matters}. We collapse it into one feed-forward pass by fixing the time embedding and removing the noise,
\begin{equation}
\mathbf{Z} \;=\; F_{\theta}\!\left(\mathcal{E}(\mathbf{x}); \, \tau \!=\! \tau_0, \, \epsilon \!=\! 0\right),
\label{eq:collapse}
\end{equation}
where $\tau_0$ is a fixed timestep and $\mathbf{Z} = \{\mathbf{z}_{ijk}\}$ is the resulting token field. With the time embedding fixed and no noise injected, $F_{\theta}$ no longer denoises; it deterministically transforms the encoded RGB field into a task-adapted token field in one pass~\cite{xu2025matters}. We fine-tune the backbone for this translation. A DiT represents the clip as a regular grid of spatio-temporal tokens,
each tied to a fixed space-time region exploited by our readout.

\subsection{Pixel Readout}
\label{sec:method:readout}

\paragraph{A correspondence the encoder already fixed.}
A DiT organizes its representations on a regular
spatio-temporal token grid, where each token
$\mathbf{z}_{ijk}$ corresponds to a fixed output
region $\Omega_{ijk}$ through patch-aligned indexing. A token is thus a spatio-temporal carrier; its channels, pretrained to hold an RGB latent, can instead hold the opacity of that block. Reading alpha is then not a reconstruction but an inversion of this existing correspondence, with the target swapped from RGB to alpha.

\paragraph{A single linear readout.}
After adaptation, each token is mapped to its block of alpha pixels by one shared linear projection:
\begin{equation}
\hat{\mathbf{A}}_{ijk} \;=\; \mathrm{reshape}\!\left(\mathbf{W}\, \mathbf{z}_{ijk} + \mathbf{b}\right), \qquad \hat{\mathbf{A}}_{ijk} \in \mathbb{R}^{\,t \times p \times p \times 1},
\label{eq:readout}
\end{equation}
where $t\!\times\!p\!\times\!p$ is exactly the pixel block $\mathbf{z}_{ijk}$ encodes. Tiling all blocks over $\{\Omega_{ijk}\}$ yields the full-resolution alpha sequence $\hat{\boldsymbol{\alpha}}$. The projection $\mathbf{W}$ is shared across all positions and contains no convolution, no learned upsampling, and no inter-token mixing.

\paragraph{A minimal token-local readout.}
We intentionally choose the simplest token-local output interface to test
whether an adapted video prior can expose alpha directly in pixel space
without a task-specific decoder. Empirically, this minimal readout already
achieves strong matting accuracy and substantially outperforms the matched
latent-output interface (Sec.~\ref{sec:ablation}). Our claim is therefore
not that richer local readouts can never help, but that a high-capacity
generative decoder is not required for accurate pixel-space alpha readout.

\textbf{Sequence-length-independent readout.} The shared readout acts independently on each spatio-temporal token and introduces no recurrence, memory bank, or dependence on sequence length. Sequence-length handling is left to the video backbone and evaluated empirically in Sec.~4.4.

\subsection{Pixel-Space Supervision}
\label{sec:method:novae}

The readout in Eq.~\eqref{eq:readout} produces alpha directly in pixel space, so the alpha target never enters the VAE. We supervise it with a pixel-space objective only:
\begin{equation}
\mathcal{L} \;=\; \big\| \hat{\boldsymbol{\alpha}} - \boldsymbol{\alpha} \big\|_{1},
\label{eq:loss}
\end{equation}
There is no alpha VAE encoding, no latent flow-matching or denoising objective, and no VAE decoding pass.

This is a structural departure from the latent interface of recent generative matting, where the ground-truth alpha is encoded by the video VAE and supervised on the alpha latent, a decoded pixel loss serving only to refine the output. That interface is constrained in two task-specific ways. \emph{(i) Target compression.} The VAE is optimized for RGB reconstruction; encoding the alpha target through it compresses the sub-pixel hair, semi-transparent boundaries, and motion-blurred edges matting must recover, capping attainable detail before learning begins. \emph{(ii) Objective placement.} A latent flow-matching or denoising loss is naturally applied in latent
space, leaving pixel-space supervision through the decoder as an indirect
refinement.

This cost is acknowledged within the latent paradigm itself: a representative generative video matting method reports that the VAE's reconstruction quality upper-bounds attainable matting accuracy, and deliberately restricts the VAE's temporal compression to avoid motion-blur artifacts when decoding alpha~\cite{ge2025generative}. Our readout removes the bound at its source by never encoding alpha.

Both costs vanish when alpha stays in pixel space. On the accuracy side, an oracle round-trip of ground-truth alpha through the frozen RGB-trained VAE already incurs MAD 3.50/1.22/0.54 on VideoMatte/V-HIM60/YouTubeMatte, losing fine hair structure before learning (Fig.~\ref{fig:latent_vs_pixel}). On the efficiency side, removing the high-resolution VAE decode pass makes our pixel readout $1.51\times$ faster than the latent interface baseline while matching or exceeding its accuracy. Thus, not encoding alpha preserves detail, while not decoding it reduces latency.
\section{Experiments}
\label{sec:Experiments}

\begin{table}[t]
\centering
\setlength{\tabcolsep}{4pt}
\renewcommand{\arraystretch}{1.15}
\caption{Public training data. We train only on existing public matting supervision and construct no additional dataset. ($\alpha$: continuous alpha; img: image; vid: video.)}
\label{tab:train_data}
\resizebox{\linewidth}{!}{%
\begin{tabular}{@{}llrll@{}}
\toprule
Dataset & Type & \multicolumn{1}{c}{Size} & GT & Source \\
\midrule
VideoMatte240K~\cite{lin2021real} & vid (greenscreen) & 479 clips & $\alpha$ & RVM, CVPR'21 \\
P3M-10K~\cite{li2021privacy}          & img (portrait)    & 9{,}421 imgs & $\alpha$ & P3M, MM'21 \\
V-HIM2K5~\cite{huynh2024maggie}      & vid (multi-inst.) & 2{,}461 clips & $\alpha$ & MaGGIe, CVPR'24 \\
\bottomrule
\end{tabular}}
\end{table}

\begin{table*}[t]
\centering
\small
\setlength{\tabcolsep}{6pt}
\renewcommand{\arraystretch}{1.1}
\begin{tabular}{l cccc | cccc}
\toprule
& \multicolumn{4}{c}{\textbf{VideoMatte} ($1920\times1080$)}
& \multicolumn{4}{c}{\textbf{YouTubeMatte} ($1920\times1080$)} \\
\cmidrule(lr){2-5}\cmidrule(lr){6-9}
Method
& MAD$\downarrow$ & MSE$\downarrow$ & Grad$\downarrow$ & dtSSD$\downarrow$
& MAD$\downarrow$ & MSE$\downarrow$ & Grad$\downarrow$ & dtSSD$\downarrow$ \\
\midrule
\multicolumn{9}{l}{\textit{Mask-guided methods}} \\
AdaM (CVPR'23)~\cite{lin2023adaptive}
& 4.42 & 0.39 & 5.12 & 1.39 & -- & -- & -- & -- \\
FTP-VM (CVPR'23)~\cite{huang2023end}
& 8.00 & 3.24 & 23.75 & 2.37 & 6.49 & 4.58 & 29.78& 2.41 \\
MaGGIe (CVPR'24)~\cite{huynh2024maggie}
& 4.42 & 0.40 & 4.03 & 1.31 & 2.37 & 0.98 & 7.69 & 1.77 \\
MatAnyone (CVPR'25)~\cite{yang2025matanyone}
& 4.24 & 0.33 & 4.00 & 1.19 & 1.99 & 0.71 & 8.91 & 1.65 \\
MatAnyone2 (CVPR'26)~\cite{yang2026matanyone}
& 4.10 & 0.28 & 3.45 & 1.15 & 1.61 & 0.50 & 7.14 & 1.53 \\
VideoMaMa$^\dagger$ (CVPR'26)~\cite{lim2026videomama}
& - & - & - & - & 1.36 & 0.29 & 3.57 & - \\
\midrule
\multicolumn{9}{l}{\textit{Auxiliary-free methods}} \\
MODNet (AAAI22)~\cite{ke2022modnet}
& 11.13 & 5.54 & 15.30 & 3.08 & 15.29 & 12.68 & 8.42 & 2.74 \\
RVM (WACV'22)~\cite{lin2022robust}
& 6.57 & 1.93 & 10.55 & 1.90 & 4.37 & 2.25 & 15.10 & 2.28 \\
GVM (SIGGRAPH'25)~\cite{ge2025generative}
& 6.33 & 2.08 & 8.04 & 1.59 & 2.68 & 1.34 & 8.38 & 1.72 \\

\midrule
\textbf{TokenMatte (Ours)}
& \best{\textbf{5.11}} & \best{\textbf{1.10}} & \best{\textbf{4.34}} & \best{\textbf{1.37}}
& \best{\textbf{1.21}} & \best{\textbf{0.25}} & \best{\textbf{3.46}} & \best{\textbf{1.42}} \\
\bottomrule
\end{tabular}

\caption{Quantitative comparison on \textbf{VideoMatte} and
\textbf{YouTubeMatte} at $1920\times1080$.
Lower is better; \textcolor{bestred}{blue} marks the best auxiliary-free results.
$^\dagger$VideoMaMa uses first-frame mask guidance with SAM2 propagation;
unreported entries are denoted by --.}

\label{tab:main_ytvm}
\end{table*}

\subsection{Experimental Setup}
\label{sec:setup}
 
\paragraph{Backbone and adaptation.} We build our model on a pretrained video DiT (Wan2.1 1.3B~\cite{wan2025wan}). The VAE encoder $\mathcal{E}$ remains frozen as the pretrained RGB input interface, while the full DiT backbone is fine-tuned for RGB-to-alpha prediction. The backbone runs in a single deterministic forward pass with the time embedding fixed to $\tau_0$ and no noise injected. The output interface remains deliberately lightweight: the only newly introduced spatial module is the shared linear token-to-pixel readout $W$.

\paragraph{Training objective and schedule.} We supervise $\hat{\alpha}$ purely in pixel space with an $\ell_1$ loss. The model is trained on 21-frame clips at $720\times1280$ resolution for 30K steps using AdamW with bfloat16 precision, taking approximately 12 hours on a TPU v6e-16. We use the same training configuration across all experiments and introduce no auxiliary trimap, mask, or segmentation supervision.

\paragraph{Inference.} At test time, we run full-sequence inference in a single forward pass, processing clips of up to $100$ frames at $1080$p at once, with no windowing or inter-segment propagation. The effect of inference length beyond the 21-frame training context is evaluated in Sec.~\ref{sec:ablation}.
 
\paragraph{Evaluation metrics.}
Following RVM~\cite{lin2022robust} and GVM~\cite{ge2025generative}, we report
MAD, MSE, Grad~\cite{rhemann2009perceptually}, Conn (connectivity), and
dtSSD~\cite{erofeev2015perceptually}; all metrics are lower-is-better.
 
\paragraph{Comparison protocol.}
We focus on the auxiliary-free setting,
where the model receives only RGB video with no trimap, mask,
point prompt, or first-frame annotation at inference.
We compare primarily against auxiliary-free methods and report
mask-guided methods for context. VideoMaMa~\cite{lim2026videomama}
is included under its first-frame mask-guided protocol, where SAM2
propagates the initial mask across the video; it is therefore not
directly comparable to our RGB-only setting. 
We follow the unified VideoMatte and YouTubeMatte evaluation protocol of MatAnyone2~\cite{yang2026matanyone}.

\begin{table*}[t]
\centering
\small
\setlength{\tabcolsep}{5pt}
\renewcommand{\arraystretch}{1.1}
\begin{tabular}{l cccc cccc cccc}
\toprule
& \multicolumn{4}{c}{\textbf{V-HIM60 (Easy)}}
& \multicolumn{4}{c}{\textbf{V-HIM60 (Medium)}}
& \multicolumn{4}{c}{\textbf{V-HIM60 (Hard)}} \\
\cmidrule(lr){2-5}\cmidrule(lr){6-9}\cmidrule(lr){10-13}
Method
& MAD & Grad & Conn & dtSSD
& MAD & Grad & Conn & dtSSD
& MAD & Grad & Conn & dtSSD \\
\midrule
RVM (WACV'22)~\cite{lin2022robust}
& 6.40 & 14.27 & 11.74 & 4.10
& 10.28 & 19.82 & 20.66 & 4.56
& 17.96 & 20.63 & 34.52 & 7.03 \\
SparseMat (CVPR'23)~\cite{sun2023ultrahigh}
& 6.51 & 13.41 & 11.71 & 4.52
& 10.08 & 16.91 & 19.79 & 5.07
& 23.48 & 20.86 & 44.64 & 9.76 \\
MaGGIe (CVPR'24)~\cite{huynh2024maggie}
& 10.48 & 13.71 & 19.99 & 5.19
& 26.16 & 19.61 & 71.29 & 7.14
& 25.68 & 21.18 & 49.78 & 11.30 \\
GVM (SIGGRAPH'25)~\cite{ge2025generative}
& 6.24 & 10.13 & 10.55 & 3.26
& 10.13 & 15.00 & 13.61 & 3.55
& 11.04 & \best{13.52} & 13.16 & \best{4.11} \\
VideoMaMa$^\dagger$ (CVPR'26)~\cite{lim2026videomama}
& 1.35 & 2.89 & -- & --
& 2.31 & 4.69 & -- & --
& 2.78 & 4.83 & -- & -- \\
\midrule
\textbf{TokenMatte (Ours)}

& \best{\textbf{3.35}} & \best{\textbf{10.00}} & \best{\textbf{5.94}} & \best{\textbf{3.23}}
& \best{\textbf{5.35}} & \best{\textbf{11.59}} & \best{\textbf{10.39}} & \best{\textbf{3.54}}
& \best{\textbf{6.09}} & \textbf{15.87} & \best{\textbf{11.42}} & \textbf{4.40} \\
\bottomrule
\end{tabular}

\caption{Quantitative comparison on \textbf{V-HIM60} across
\textit{Easy}, \textit{Medium}, and \textit{Hard} splits.
Lower is better; \textcolor{bestred}{blue} marks the best auxiliary-free results.
$^\dagger$VideoMaMa uses first-frame mask guidance with SAM2 propagation;
Conn and dtSSD are not reported.}

\label{tab:main_vhim}
\end{table*}

\subsection{Datasets}
\label{sec:datasets}
 
\paragraph{Training data.} Public alpha supervision remains limited in scale and diversity: existing video matting datasets are largely composited or green-screen captured and dominated by human portraits. We therefore adapt a pretrained video DiT using existing public alpha supervision rather than constructing an additional task-specific corpus. Concretely, we use VideoMatte240K~\cite{lin2021real} as the primary video matte source, P3M-10K~\cite{li2021privacy} for large-scale portrait fine-alpha supervision, and V-HIM2K5~\cite{huynh2024maggie} for multi-instance video matting; their statistics are summarized in Table~\ref{tab:train_data}. This differs from recent data-scaling lines of work that construct additional task-specific resources---e.g.\ GVM~\cite{ge2025generative} renders the synthetic SynHairMan set and pseudo-labels real video segmentation, VideoMaMa~\cite{lim2026videomama} builds the 50K-video pseudo-labeled MA-V dataset via a generative mask-to-matte labeler, and MatAnyone\,2~\cite{yang2026matanyone} curates the real-world VMReal dataset ($28$K clips, $2.4$M frames) with a learned quality evaluator. TokenMatte instead uses only the public datasets listed above while leveraging pretrained video representations for task adaptation.

\paragraph{Evaluation benchmarks.}
We evaluate on three standard public benchmarks, all in the
auxiliary-free regime with no benchmark-specific fine-tuning.
\textbf{VideoMatte}~\cite{lin2021real} provides $50$ composited test clips
($25$ motion $+$ $25$ static) of $100$ frames each at $1920\times1080$,
following the RVM~\cite{lin2022robust} evaluation split with continuous-$\alpha$
ground truth.
\textbf{YouTubeMatte}~\cite{yang2025matanyone} is a high-resolution
($1920\times1080$) human video matting benchmark of $64$ clips
($32$ motion $+$ $32$ static), $100$ frames each, harmonized from raw
green-screen footage.
\textbf{V-HIM60}~\cite{huynh2024maggie} is a multi-instance human matting
benchmark of $60$ clips split evenly into Easy / Medium / Hard
($20$ each), $30$ frames per clip, with manually annotated
continuous-$\alpha$ ground truth that preserves the finest hair-level
fly-aways among the three benchmarks.

\subsection{Main Results}
\label{sec:main_results}

\paragraph{Comparison with prior work.}
Tables~\ref{tab:main_ytvm} and~\ref{tab:main_vhim} report results on the
three benchmarks. Across all of them our method sets a new
auxiliary-free state of the art, and the margin widens precisely on the
cases that stress real-world appearance and multi-instance structure.
On the composited \textbf{VideoMatte} benchmark we lead every
auxiliary-free baseline on all four metrics---e.g.\ MAD $5.11$ vs.\
$6.33$ and Grad $4.34$ vs.\ $8.04$ for GVM, nearly halving the gradient
error---indicating substantially sharper boundaries. On YouTubeMatte the advantage is far larger: at MAD 1.21,
MSE 0.25, and Grad 3.46 we not only top the auxiliary-free
group but also outperform the mask-guided VideoMaMa
(1.36/0.29/3.57) on all three commonly reported metrics,
despite using no auxiliary input. These results are obtained using only the public training datasets described in Sec.~\ref{sec:datasets}, without constructing an additional task-specific matting corpus. The trend is most pronounced on \textbf{V-HIM60} (Table~\ref{tab:main_vhim}), where multiple interacting human instances are the main difficulty. Among auxiliary-free methods, TokenMatte achieves the best MAD and Conn on all three splits and the best Grad and dtSSD on Easy and Medium, with especially large gains in MAD on the hardest split (6.09 vs.\ 11.04 for GVM). On Hard, GVM obtains lower Grad (13.52 vs.\ 15.87) and dtSSD (4.11 vs.\ 4.40), while TokenMatte retains lower MAD (6.09 vs.\ 11.04) and Conn (11.42 vs.\ 13.16). Mask-guided methods are reported only for context under their stronger input conditions.

\subsection{Ablation Studies}
\label{sec:ablation}

\paragraph{Latent baseline.}
To isolate the output interface, we compare against a controlled \emph{Latent} baseline with the same Wan2.1 backbone, training data, and configuration. Instead of the per-token pixel readout, it encodes ground-truth alpha with the video VAE, supervises the alpha latent, and decodes predictions with a frozen VAE decoder at inference, following recent generative matting~\cite{ge2025generative,lim2026videomama}. Thus, the comparison changes only the alpha supervision and output interface. Latency is measured on a single $100$-frame $1920\times1080$ clip using one NVIDIA A100 in eager mode; Fig.~\ref{fig:latent_vs_pixel}(d) attributes the gap to the VAE decoding stage.

\begin{table}[h]
\centering
\small
\setlength{\tabcolsep}{4pt}
\renewcommand{\arraystretch}{1.1}
\begin{tabular}{l ccccc}
\toprule
Paradigm
& MAD$\downarrow$ & MSE$\downarrow$ & Grad$\downarrow$ & dtSSD$\downarrow$ & Latency$\downarrow$ \\
\midrule
Random Init
& 21.5 & 16.0 & 61.0 & 5.46 & 382ms\\                                                                  
LATENT
& 6.00 & 1.36 & 8.22 & 1.89 & 576ms\\
PIXEL (Ours)
& \best{\textbf{5.11}} & \best{\textbf{1.10}} & \best{\textbf{4.34}} & \best{\textbf{1.37}} & \best{\textbf{382ms}} \\
\bottomrule
\end{tabular}
\caption{\textbf{Paradigm ablation.} The large gap from Random Init confirms the importance of the pretrained video prior, while the improvement from LATENT to PIXEL shows that direct pixel-space alpha readout preserves fine matte details and avoids VAE decoding overhead.}
\vspace{-0.3cm}
\label{tab:ablation_paradigm}
\end{table}

Table~\ref{tab:ablation_paradigm} isolates the two design choices behind our framework: the pretrained video prior and the pixel-space readout. Removing the pretrained initialization (\emph{Random Init}) substantially degrades performance across every metric (MAD $21.5$, Grad $61.0$), confirming the importance of the pretrained video prior under the same data and optimization setting. The gap between the latent and pixel interfaces is especially pronounced on Grad, which improves from $8.22$ to $4.34$ with the pixel-native formulation. Since Grad is particularly sensitive to fine boundary structures, this result is consistent with the effect of the output interface: encoding the alpha target through an RGB-optimized VAE compresses the sub-pixel hair and semi-transparent edges that matting must recover, whereas reading alpha directly off the token grid preserves these local details more faithfully. The pixel interface also improves MAD, MSE, and dtSSD over its latent counterpart. Skipping VAE decoding reduces per-clip latency from $576$\,ms to $382$\,ms, yielding a $1.51\times$ speedup. Together, these ablations demonstrate the importance of the pretrained video prior and show that direct pixel-space alpha readout provides a more accurate and efficient interface than latent supervision.

\paragraph{Readout design.} We further test whether the minimal linear readout limits prediction quality by replacing it with a shallow local convolutional decoder while keeping the backbone, training data, and optimization unchanged. As shown in Table~\ref{tab:readout_design}, the convolutional decoder changes MAD only marginally from 5.11 to 5.09 while slightly worsening Grad from 4.34 to 4.38. Thus, additional local decoding provides no consistent improvement, supporting the use of the shared linear readout as a simple and effective pixel-native output interface.

\begin{table}[h]
\centering
\small
\setlength{\tabcolsep}{7pt}
\renewcommand{\arraystretch}{1.08}
\begin{tabular}{lcc}
\toprule
Readout & MAD$\downarrow$ & Grad$\downarrow$ \\
\midrule
Linear (Ours) & 5.11 & \best{\textbf{4.34}} \\
Local convolutional decoder & \best{\textbf{5.09}} & 4.38 \\
\bottomrule
\end{tabular}
\caption{\textbf{Readout design on VideoMatte.} A shallow local convolutional decoder provides no consistent improvement over the shared linear readout.}
\vspace{-0.3cm}
\label{tab:readout_design}
\end{table}

\paragraph{Adaptation.}
We further examine whether the pixel-native formulation depends on a
particular backbone-adaptation regime. We compare our default backbone
adaptation with a parameter-efficient LoRA variant while keeping the same
pixel-native output interface. As shown in
Table~\ref{tab:backbone_adaptation}, LoRA already transfers strongly to
VideoMatte, even obtaining a slightly lower MAD, but its performance
degrades substantially on the held-out YouTubeMatte benchmark. This
suggests that the pretrained video prior provides a strong starting point,
while stronger task adaptation remains important for robust cross-dataset
transfer. We therefore use a single fixed backbone-adaptation configuration
across all benchmarks rather than selecting the adaptation strategy
separately for each test set.

\begin{table}[h]
\centering
\small
\setlength{\tabcolsep}{7pt}
\renewcommand{\arraystretch}{1.1}
\begin{tabular}{lcc}
\toprule
Adaptation
& VideoMatte $\downarrow$
& YouTubeMatte $\downarrow$ \\
\midrule
LoRA
& \best{\textbf{4.11}} & 16.4 \\
Full (Ours)
& 5.11 & \best{\textbf{1.21}} \\
\bottomrule
\end{tabular}
\caption{\textbf{Backbone adaptation.}
MAD on VideoMatte and YouTubeMatte under LoRA and full adaptation. Full adaptation substantially improves cross-dataset generalization, while both use the same pixel-native output interface.}
\vspace{-0.3cm}
\label{tab:backbone_adaptation}
\end{table}

\paragraph{Inference length.} Table~\ref{tab:sequence_length} evaluates the same checkpoint using inference contexts of 21, 61, and 100 frames, while training uses 21-frame clips. MAD, MSE, and Grad remain stable or improve as the inference context increases, with MAD decreasing from 6.04 to 5.11 and Grad from 4.63 to 4.34. Temporal consistency is also largely preserved, as dtSSD changes only marginally from 1.33 to 1.37. These results indicate that the model remains stable when evaluated substantially beyond its training clip length, including full-sequence 100-frame inference without windowing or inter-segment propagation.

\begin{table}[h]
\centering
\small
\setlength{\tabcolsep}{6pt}
\renewcommand{\arraystretch}{1.08}
\begin{tabular}{lcccc}
\toprule
Inference length
& MAD$\downarrow$
& MSE$\downarrow$
& Grad$\downarrow$
& dtSSD$\downarrow$ \\
\midrule
21
& 6.04 & 1.17 & 4.63 & \best{\textbf{1.33}} \\
61
& 5.59 & 1.13 & 4.47 & 1.35 \\
100$^*$
& \best{\textbf{5.11}}
& \best{\textbf{1.10}}
& \best{\textbf{4.34}}
& 1.37 \\
\bottomrule
\end{tabular}
\caption{\textbf{Inference length on VideoMatte.}
The same 21-frame-trained checkpoint is evaluated on complete 100-frame
videos with different inference lengths. Spatial metrics remain stable or
improve, while dtSSD changes only marginally.
$^*$The full sequence is padded to 101 frames for Wan2.1's $4k+1$
requirement; padding is excluded from evaluation.}
\vspace{-0.3cm}
\label{tab:sequence_length}
\end{table}

\begin{figure*}[h]
    \centering
    \includegraphics[width=\textwidth]{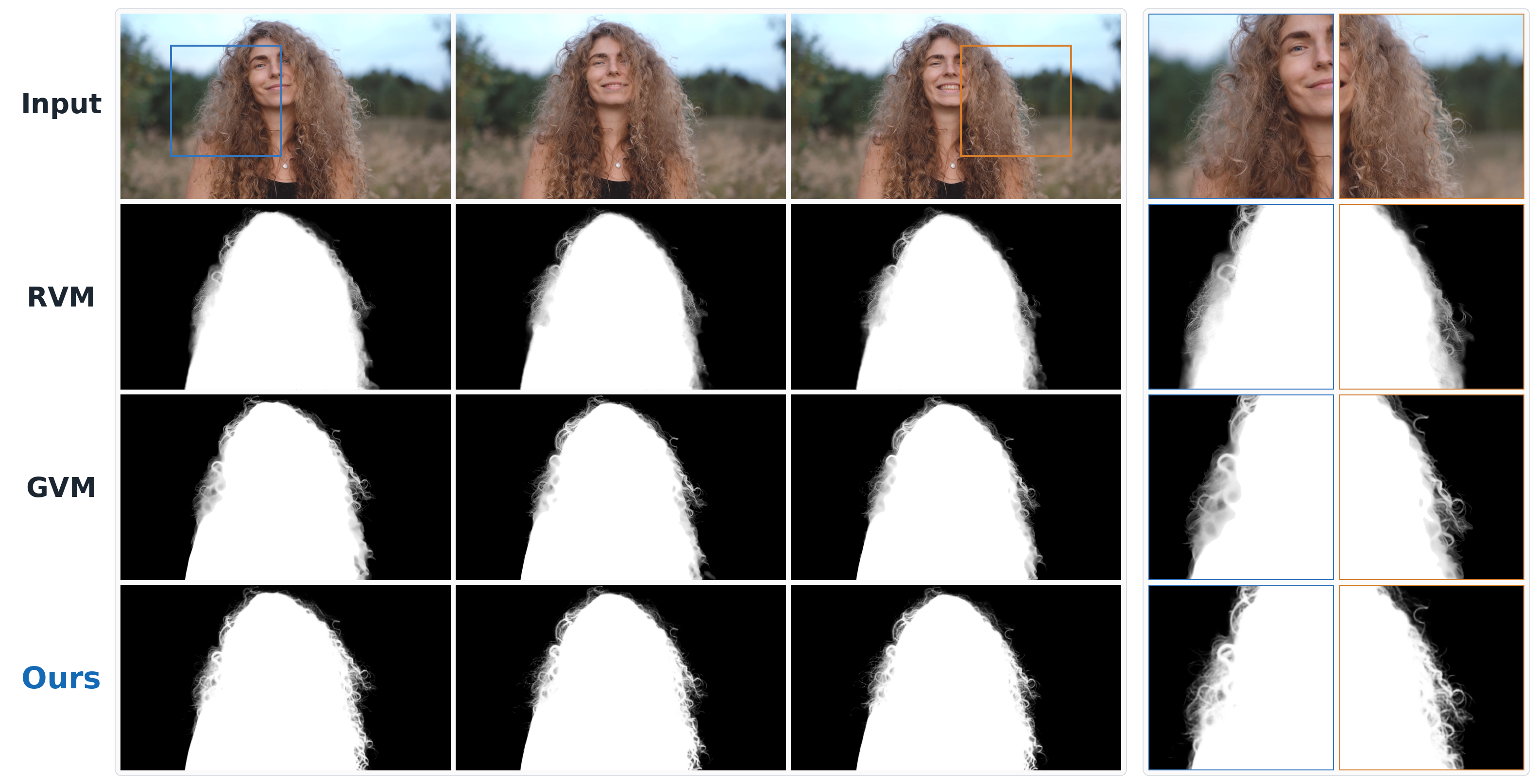}

\caption{\textbf{Qualitative comparison with auxiliary-free methods.} We compare TokenMatte with RVM and latent-based GVM across three video frames. The blue and orange boxes indicate regions magnified on the right. RVM produces diffuse alpha boundaries, while GVM exhibits over-smoothed predictions that blur fine structures and merge neighboring details. In contrast, TokenMatte preserves sharper boundaries and finer, more separated alpha structures without auxiliary inputs.}

    \label{fig:qualitative}
\end{figure*}

\subsection{Qualitative Results}
\begin{figure}[t]
\centering
\includegraphics[width=\linewidth]{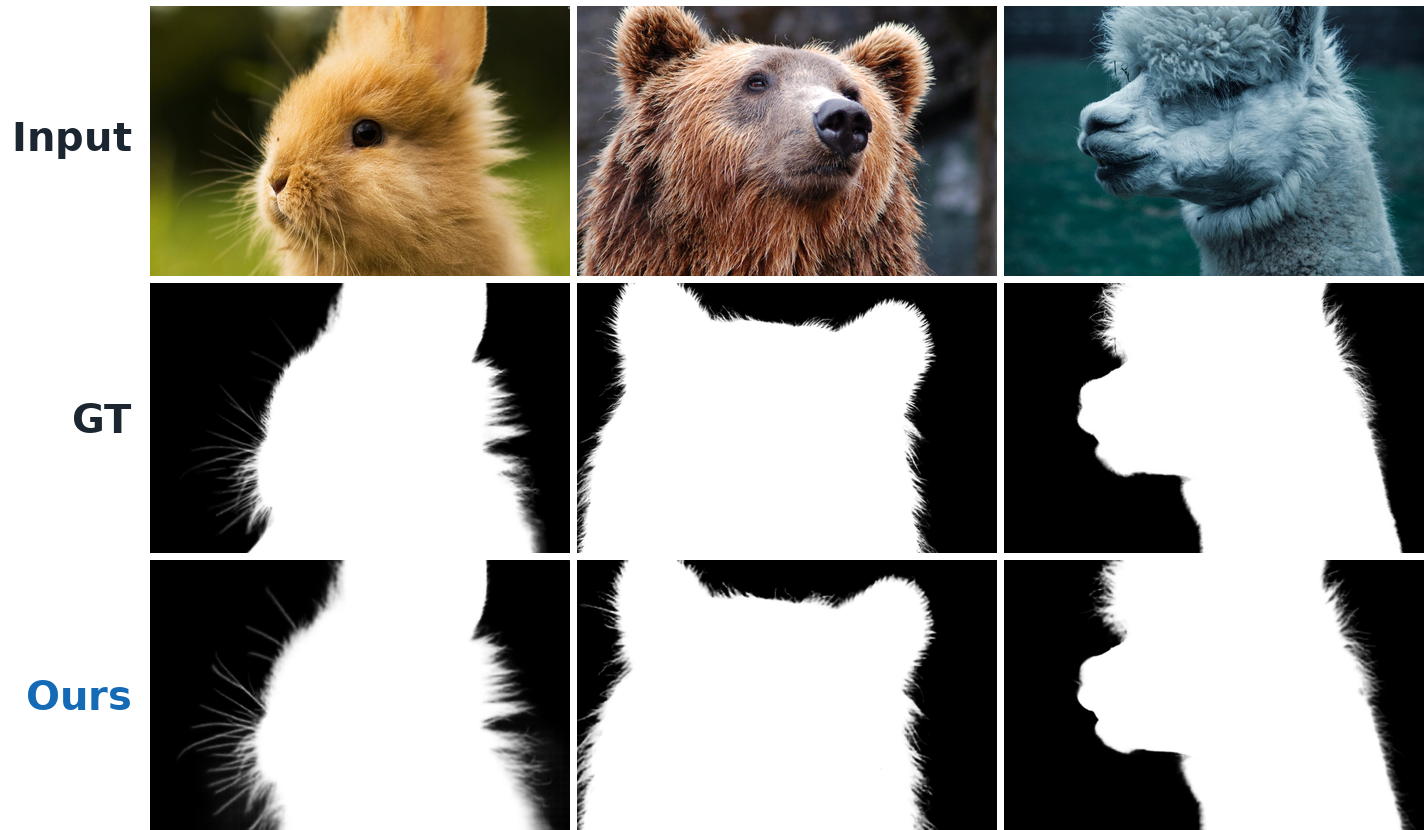}
\caption{\textbf{Generalization to non-human foregrounds.} TokenMatte produces detailed alpha mattes across diverse animal categories, preserving fine whiskers, fur, and irregular boundary structures beyond the predominantly human-centered evaluation domain. Ground-truth mattes are shown for reference.}
\vspace{-0.2cm}
\label{fig:animal_generalization}
\end{figure}

Figure~\ref{fig:qualitative} compares TokenMatte with RVM and GVM across three frames of a challenging curly-hair sequence. In the magnified regions, RVM produces diffuse alpha responses, while GVM over-smooths curl boundaries and merges neighboring strands into broader structures; TokenMatte preserves sharper and more separated hair details across the shown frames. Figure~\ref{fig:animal_generalization} further demonstrates qualitative generalization beyond the predominantly human-centered benchmarks, where TokenMatte recovers fine whiskers, fur, and irregular boundary structures across diverse animal foregrounds.

\section{Conclusion}
We presented TokenMatte, an auxiliary-free video matting framework that adapts a pretrained video DiT with a lightweight token-to-pixel output interface. A shared linear readout maps each adapted spatio-temporal token directly to its corresponding block of alpha pixels, keeping alpha supervision entirely in pixel space. This design avoids the information loss introduced by encoding continuous alpha through an RGB-trained VAE and eliminates the VAE decoding stage at inference. Under the same backbone, training data, and optimization setting, the pixel-native interface achieves better matting accuracy and $1.51\times$ faster inference than its matched latent counterpart. Our ablations further confirm the importance of both the pretrained video prior and the direct pixel-space output interface for accurate alpha prediction. Using only public alpha supervision without constructing an additional matting corpus, TokenMatte establishes a new auxiliary-free state of the art on VideoMatte, YouTubeMatte, and V-HIM60. These results demonstrate that pretrained video DiT representations can be effectively adapted for high-resolution video matting through a simple pixel-native readout, enabling accurate continuous alpha prediction without a heavy task-specific decoder.

\section{Additional Results and Details}
\label{sec:additional}
\subsection{Relation to Prior Generative Matting}
\label{sec:supp_relation}
For completeness, Table~\ref{tab:supp_relation} contrasts TokenMatte with recent generative and data-scaling video matting methods at the design level. Latent approaches reach the alpha target through the backbone's VAE---encoding alpha into an RGB-trained latent, supervising there, and decoding back---and several pair this with large self-curated corpora or auxiliary mask guidance. We instead keep alpha entirely in pixel space and use only public supervision, so our gains come from unlocking the pretrained video prior rather than from latent supervision, extra curated data, or auxiliary inputs.
\begin{table}[!h]
\centering
\small
\setlength{\tabcolsep}{3.6pt}
\renewcommand{\arraystretch}{1.10}
\begin{tabular}{lcccc}
\toprule
\textbf{Method} & \textbf{Alpha} & \makecell{\textbf{Matting}\\\textbf{data}} & \textbf{Guidance} & \makecell{\textbf{VAE}\\\textbf{dec.}} \\
\midrule
GVM & latent & synth.+pseudo & none & yes \\
VideoMaMa & latent & synthetic only & mask & yes \\
MatAnyone2 & pixel & VMReal (28K) & mask & -- \\
\rowcolor{bestred!25}
\textbf{Ours} & pixel & public only & none & no \\
\bottomrule
\end{tabular}
\caption{\textbf{Design-level comparison.} TokenMatte uses pixel-space alpha, no auxiliary guidance, and no VAE decoding.}
\label{tab:supp_relation}
\end{table}
\subsection{Implementation Details}
\label{sec:supp_impl}
We use the same fixed timestep $\tau_0$ during training and inference with no injected noise ($\epsilon=0$). The VAE encoder remains frozen, while the full video DiT is fine-tuned using the public training datasets described in the main paper. Additional optimization and inference configurations are summarized in Table~\ref{tab:supp_impl}.
\begin{table}[!t]
\centering
\setlength{\tabcolsep}{4pt}
\begin{tabular}{ll}
\toprule
\textbf{Configuration} & \textbf{Value} \\
\midrule
\multicolumn{2}{l}{\emph{Optimization}}\\
Optimizer & AdamW ($\beta_1{=}0.9,\ \beta_2{=}0.95$)\\
Learning rate & $2\times10^{-5}$, constant \\
Weight decay & $1\times10^{-3}$ \\
Gradient clipping & global norm $1.0$ \\
Global batch size & $64$ \\
Training steps & $30\text{K}$ \\
Precision & bfloat16 \\
\midrule
\multicolumn{2}{l}{\emph{Forward pass}}\\
Fixed timestep & $\tau_0$ (no noise, $\epsilon{=}0$) \\
Readout head & single shared linear \\
Training loss & $\ell_1$ in pixel space (Eq.~3)\\
\midrule
\multicolumn{2}{l}{\emph{Data / augmentation}}\\
Training clip length & $21$ frames \\
Training resolution & $720\times1280$ \\
Augmentation & horizontal flip \\
Background pool & RVM background composites \\
\midrule
\multicolumn{2}{l}{\emph{Inference}}\\
Mode & full-sequence single pass\\
Clip length & up to $100$ frames \\
Resolution & $1920\times1080$ \\
\bottomrule
\end{tabular}
\caption{Full training and inference configuration of TokenMatte.}
\label{tab:supp_impl}
\end{table}
\subsection{Additional Qualitative Results}
\label{sec:supp_qualitative}
\begin{figure*}[!t]
\centering
\includegraphics[width=\textwidth]{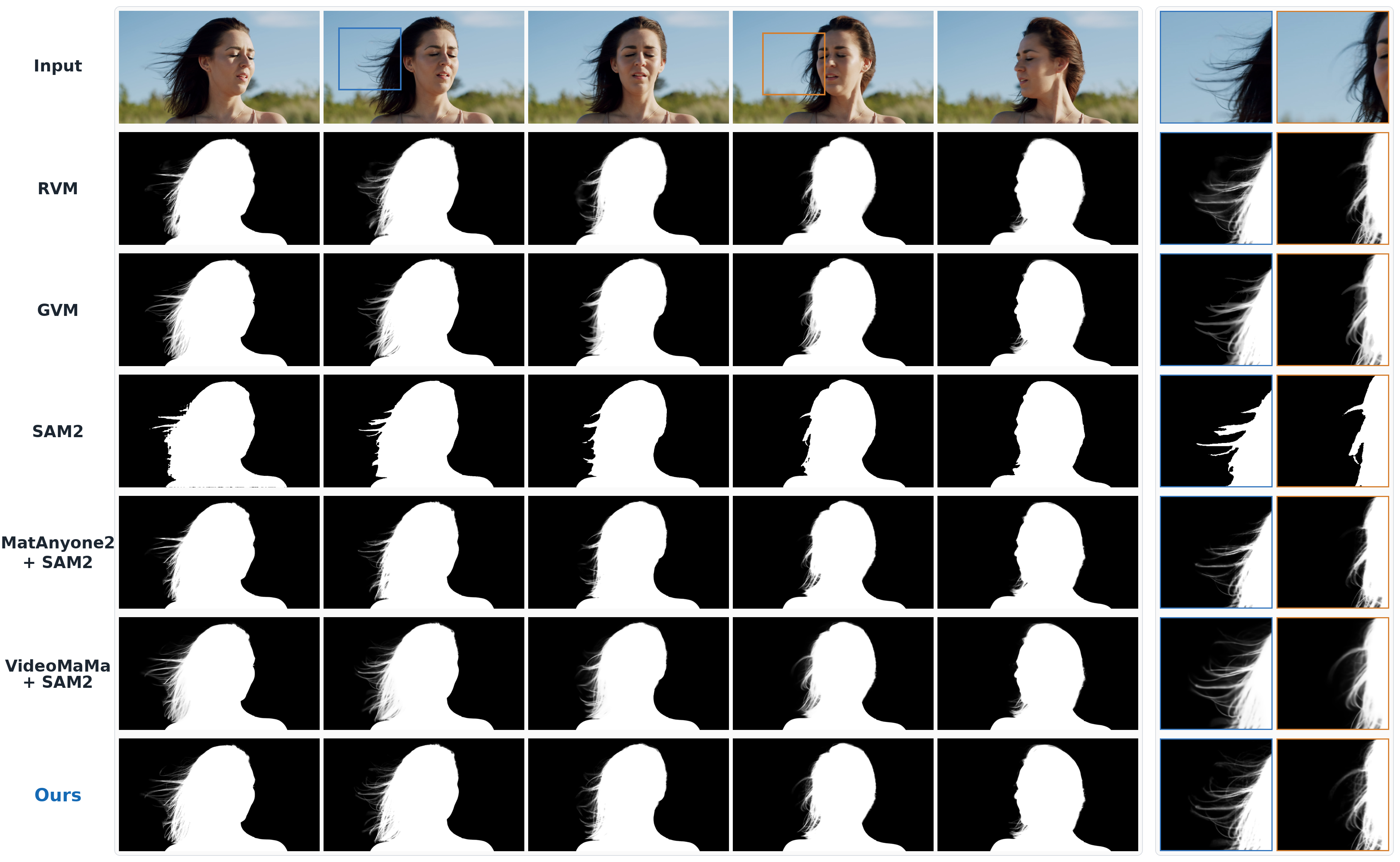}
\caption{\textbf{Qualitative comparison with auxiliary-free and mask-guided methods.} We compare TokenMatte with auxiliary-free RVM and GVM, and with mask-guided MatAnyone2 and VideoMaMa using SAM2 guidance. The SAM2 mask sequence is shown for reference, and colored boxes indicate the magnified regions on the right. While all methods recover the main foreground, RVM produces diffuse hair boundaries and the other baselines tend to smooth some thin strands. TokenMatte preserves finer and more separated hair structures without auxiliary mask guidance.}
\label{fig:supp_comparison}
\end{figure*}

\begin{figure*}[!t]
\centering
\includegraphics[width=\textwidth]{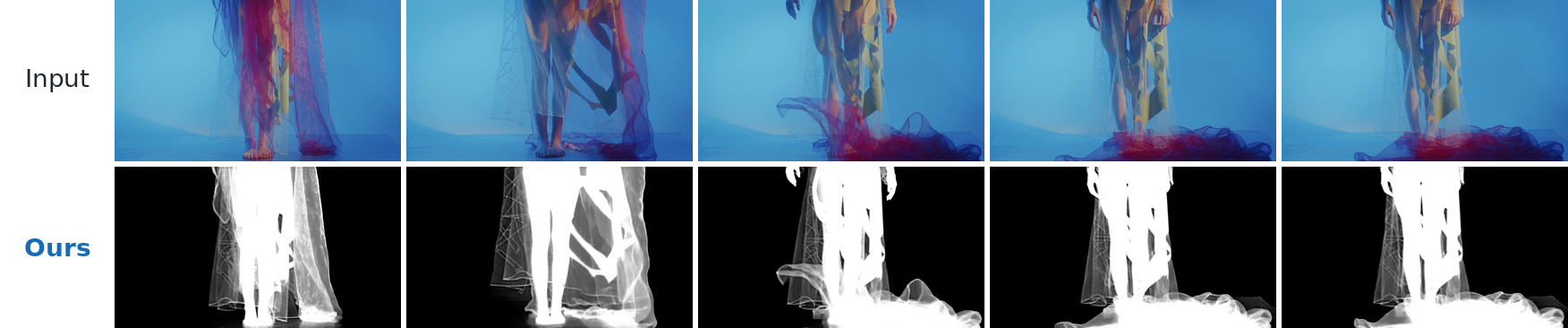}
\caption{\textbf{Translucent fabric.} TokenMatte preserves thin fabric layers and spatially varying opacity across the sequence.}
\label{fig:supp_fabric}
\end{figure*}

\begin{figure*}[!t]
\centering
\includegraphics[width=\textwidth]{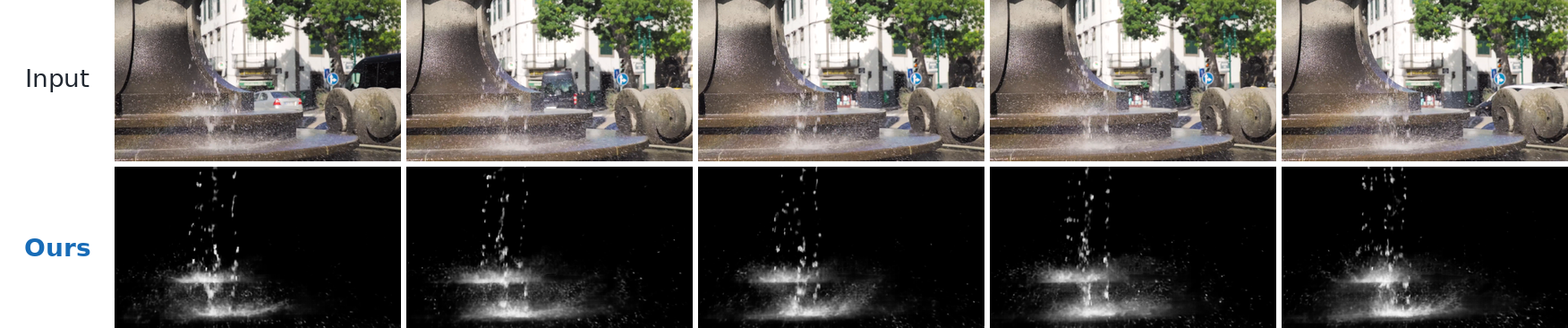}
\caption{\textbf{Fast water spray.} TokenMatte captures rapidly changing water structures and fine spray details from RGB input alone.}
\label{fig:supp_water}
\end{figure*}

\begin{figure*}[!t]
\centering
\includegraphics[width=\textwidth]{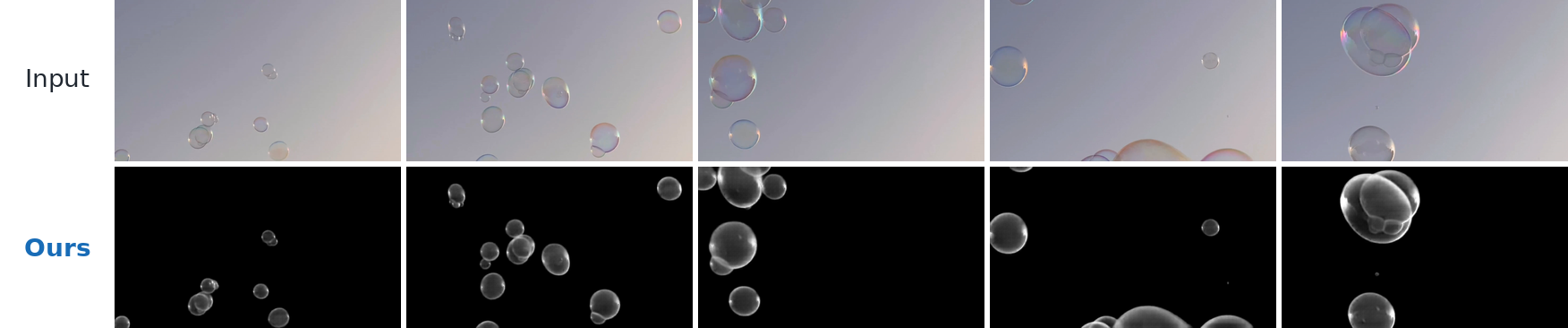}
\caption{\textbf{Transparent soap bubbles.} TokenMatte recovers transparent bubbles with clear boundaries and varying opacity over time.}
\label{fig:supp_bubbles}
\end{figure*}

\begin{figure*}[!t]
\centering
\includegraphics[width=\textwidth]{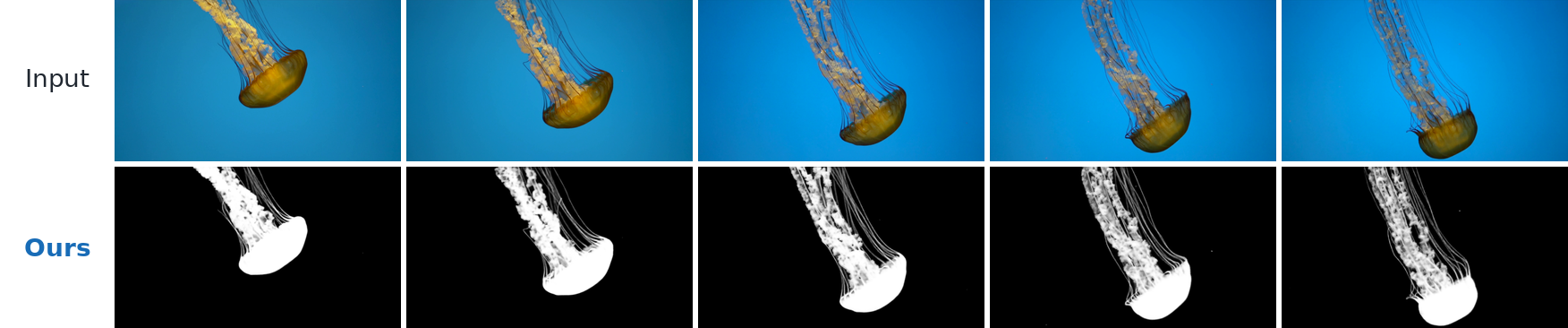}
\caption{\textbf{Translucent tentacles.} TokenMatte preserves thin tentacles and translucent structures across separated frames.}
\label{fig:supp_tentacles}
\end{figure*}

\begin{figure*}[!t]
\centering
\includegraphics[width=\textwidth]{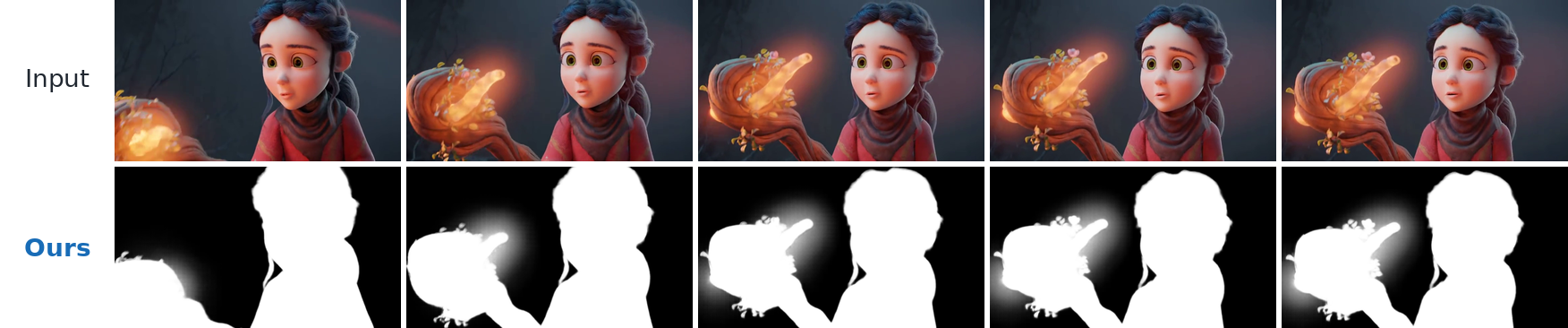}
\caption{\textbf{Soft emissive boundaries.} TokenMatte captures the foreground together with soft, spatially varying alpha transitions around the glowing hand across frames.}
\label{fig:supp_glow}
\end{figure*}

\begin{figure*}[!t]
\centering
\includegraphics[width=\textwidth]{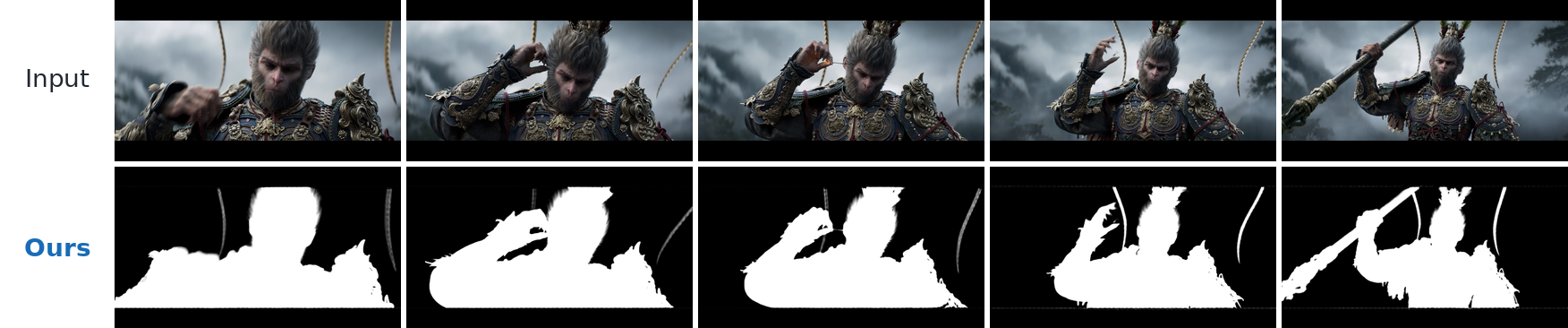}
\caption{\textbf{Complex foreground interaction.} TokenMatte preserves the articulated foreground together with thin structures such as hair, ropes, and the moving weapon across temporally separated frames.}
\label{fig:supp_wukong}
\end{figure*}

Figure~\ref{fig:supp_comparison} compares TokenMatte with auxiliary-free and mask-guided video matting methods on a challenging zero-shot sequence with fine moving hair. Figures~\ref{fig:supp_fabric}--\ref{fig:supp_wukong} further demonstrate zero-shot generalization to diverse dynamic, translucent, and non-rigid foregrounds.
\clearpage
{
\small
\bibliographystyle{ieeenat_fullname}
\bibliography{main}

@String(CVPR= {IEEE Conf. Comput. Vis. Pattern Recog.})

@String(BMVC= {Brit. Mach. Vis. Conf.})

@String(AAAI = {AAAI})

@String(CVPR  = {CVPR})

@String(BMVC  =	{BMVC})

@inproceedings{chuang2002video,
  title={Video matting of complex scenes},
  author={Chuang, Yung-Yu and Agarwala, Aseem and Curless, Brian and Salesin, David H and Szeliski, Richard},
  booktitle={Proceedings of the 29th annual conference on Computer graphics and interactive techniques},
  pages={243--248},
  year={2002}
}

@inproceedings{apostoloff2004bayesian,
  title={Bayesian video matting using learnt image priors},
  author={Apostoloff, Nicholas and Fitzgibbon, Andrew},
  booktitle={Proceedings of the 2004 IEEE Computer Society Conference on Computer Vision and Pattern Recognition, 2004. CVPR 2004.},
  volume={1},
  pages={I--I},
  year={2004},
  organization={IEEE}
}

@inproceedings{sengupta2020background,
  title={Background matting: The world is your green screen},
  author={Sengupta, Soumyadip and Jayaram, Vivek and Curless, Brian and Seitz, Steven M and Kemelmacher-Shlizerman, Ira},
  booktitle={Proceedings of the IEEE/CVF Conference on Computer Vision and Pattern Recognition},
  pages={2291--2300},
  year={2020}
}

@inproceedings{lin2022robust,
  title={Robust high-resolution video matting with temporal guidance},
  author={Lin, Shanchuan and Yang, Linjie and Saleemi, Imran and Sengupta, Soumyadip},
  booktitle={Proceedings of the IEEE/CVF Winter Conference on Applications of Computer Vision},
  pages={238--247},
  year={2022}
}

@inproceedings{ke2022modnet,
  title={Modnet: Real-time trimap-free portrait matting via objective decomposition},
  author={Ke, Zhanghan and Sun, Jiayu and Li, Kaican and Yan, Qiong and Lau, Rynson WH},
  booktitle={Proceedings of the AAAI conference on artificial intelligence},
  volume={36},
  number={1},
  pages={1140--1147},
  year={2022}
}

@inproceedings{xu2017deep,
  title={Deep image matting},
  author={Xu, Ning and Price, Brian and Cohen, Scott and Huang, Thomas},
  booktitle={Proceedings of the IEEE conference on computer vision and pattern recognition},
  pages={2970--2979},
  year={2017}
}

@inproceedings{yang2026matanyone,
  title={MatAnyone 2: Scaling video matting via a learned quality evaluator},
  author={Yang, Peiqing and Zhou, Shangchen and Hao, Kai and Tao, Qingyi},
  booktitle={Proceedings of the IEEE/CVF Conference on Computer Vision and Pattern Recognition},
  pages={37476--37485},
  year={2026}
}

@inproceedings{lin2021real,
  title={Real-time high-resolution background matting},
  author={Lin, Shanchuan and Ryabtsev, Andrey and Sengupta, Soumyadip and Curless, Brian L and Seitz, Steven M and Kemelmacher-Shlizerman, Ira},
  booktitle={Proceedings of the IEEE/CVF Conference on Computer Vision and Pattern Recognition},
  pages={8762--8771},
  year={2021}
}

@inproceedings{wang2021video,
  title={Video matting via consistency-regularized graph neural networks},
  author={Wang, Tiantian and Liu, Sifei and Tian, Yapeng and Li, Kai and Yang, Ming-Hsuan},
  booktitle={Proceedings of the IEEE/CVF International Conference on Computer Vision},
  pages={4902--4911},
  year={2021}
}

@inproceedings{huynh2024maggie,
  title={Maggie: Masked guided gradual human instance matting},
  author={Huynh, Chuong and Oh, Seoung Wug and Shrivastava, Abhinav and Lee, Joon-Young},
  booktitle={Proceedings of the IEEE/CVF Conference on Computer Vision and Pattern Recognition},
  pages={3870--3879},
  year={2024}
}

@inproceedings{cai2019disentangled,
  title={Disentangled image matting},
  author={Cai, Shaofan and Zhang, Xiaoshuai and Fan, Haoqiang and Huang, Haibin and Liu, Jiangyu and Liu, Jiaming and Liu, Jiaying and Wang, Jue and Sun, Jian},
  booktitle={Proceedings of the IEEE/CVF International Conference on Computer Vision},
  pages={8819--8828},
  year={2019}
}

@inproceedings{yang2025matanyone,
  title={MatAnyone: Stable video matting with consistent memory propagation},
  author={Yang, Peiqing and Zhou, Shangchen and Zhao, Jixin and Tao, Qingyi and Loy, Chen Change},
  booktitle={Proceedings of the Computer Vision and Pattern Recognition Conference},
  pages={7299--7308},
  year={2025}
}

@article{lim2026videomama,
  title={VideoMaMa: Mask-Guided Video Matting via Generative Prior},
  author={Lim, Sangbeom and Oh, Seoung Wug and Huang, Jiahui and Yoon, Heeji and Kim, Seungryong and Lee, Joon-Young},
  journal={arXiv preprint arXiv:2601.14255},
  year={2026}
}

@inproceedings{ge2025generative,
  title={Generative video matting},
  author={Ge, Yongtao and Xie, Kangyang and Xu, Guangkai and Ke, Li and Liu, Mingyu and Huang, Longtao and Xue, Hui and Chen, Hao and Shen, Chunhua},
  booktitle={Proceedings of the Special Interest Group on Computer Graphics and Interactive Techniques Conference Conference Papers},
  pages={1--10},
  year={2025}
}

@inproceedings{xu2025matters,
  title={What matters when repurposing diffusion models for general dense perception tasks?},
  author={Xu, Guangkai and Liu, Mingyu and Fan, Chengxiang and Xie, Kangyang and Zhao, Zhiyue and Chen, Hao and Shen, Chunhua and others},
  booktitle={International Conference on Learning Representations},
  volume={2025},
  pages={6786--6799},
  year={2025}
}

@article{wang2025deforming,
  title={Deforming Videos to Masks: Flow Matching for Referring Video Segmentation},
  author={Wang, Zanyi and Jiang, Dengyang and Li, Liuzhuozheng and Dang, Sizhe and Li, Chengzu and Yang, Harry and Dai, Guang and Wang, Mengmeng and Wang, Jingdong},
  journal={arXiv preprint arXiv:2510.06139},
  year={2025}
}

@article{wan2025wan,
  title={Wan: Open and advanced large-scale video generative models},
  author={Wan, Team and Wang, Ang and Ai, Baole and Wen, Bin and Mao, Chaojie and Xie, Chen-Wei and Chen, Di and Yu, Feiwu and Zhao, Haiming and Yang, Jianxiao and others},
  journal={arXiv preprint arXiv:2503.20314},
  year={2025}
}

@inproceedings{chen2025video,
  title={Video depth anything: Consistent depth estimation for super-long videos},
  author={Chen, Sili and Guo, Hengkai and Zhu, Shengnan and Zhang, Feihu and Huang, Zilong and Feng, Jiashi and Kang, Bingyi},
  booktitle={Proceedings of the Computer Vision and Pattern Recognition Conference},
  pages={22831--22840},
  year={2025}
}

@inproceedings{hou2019context,
  title={Context-aware image matting for simultaneous foreground and alpha estimation},
  author={Hou, Qiqi and Liu, Feng},
  booktitle={Proceedings of the IEEE/CVF International Conference on Computer Vision},
  pages={4130--4139},
  year={2019}
}

@inproceedings{li2020natural,
  title={Natural image matting via guided contextual attention},
  author={Li, Yaoyi and Lu, Hongtao},
  booktitle={Proceedings of the AAAI conference on artificial intelligence},
  volume={34},
  number={07},
  pages={11450--11457},
  year={2020}
}

@inproceedings{park2022matteformer,
  title={Matteformer: Transformer-based image matting via prior-tokens},
  author={Park, GyuTae and Son, SungJoon and Yoo, JaeYoung and Kim, SeHo and Kwak, Nojun},
  booktitle={Proceedings of the IEEE/CVF Conference on Computer Vision and Pattern Recognition},
  pages={11696--11706},
  year={2022}
}

@inproceedings{lu2019indices,
  title={Indices matter: Learning to index for deep image matting},
  author={Lu, Hao and Dai, Yutong and Shen, Chunhua and Xu, Songcen},
  booktitle={Proceedings of the IEEE/CVF International Conference on Computer Vision},
  pages={3266--3275},
  year={2019}
}

@article{li2021deep,
  title={Deep automatic natural image matting},
  author={Li, Jizhizi and Zhang, Jing and Tao, Dacheng},
  journal={arXiv preprint arXiv:2107.07235},
  year={2021}
}

@inproceedings{li2021privacy,
  title={Privacy-preserving portrait matting},
  author={Li, Jizhizi and Ma, Sihan and Zhang, Jing and Tao, Dacheng},
  booktitle={Proceedings of the 29th ACM international conference on multimedia},
  pages={3501--3509},
  year={2021}
}

@inproceedings{kirillov2023segment,
  title={Segment anything},
  author={Kirillov, Alexander and Mintun, Eric and Ravi, Nikhila and Mao, Hanzi and Rolland, Chloe and Gustafson, Laura and Xiao, Tete and Whitehead, Spencer and Berg, Alexander C and Lo, Wan-Yen and others},
  booktitle={Proceedings of the IEEE/CVF international conference on computer vision},
  pages={4015--4026},
  year={2023}
}

@inproceedings{zhang2021attention,
  title={Attention-guided temporally coherent video object matting},
  author={Zhang, Yunke and Wang, Chi and Cui, Miaomiao and Ren, Peiran and Xie, Xuansong and Hua, Xian-Sheng and Bao, Hujun and Huang, Qixing and Xu, Weiwei},
  booktitle={Proceedings of the 29th ACM International Conference on Multimedia},
  pages={5128--5137},
  year={2021}
}

@inproceedings{yu2021mask,
  title={Mask guided matting via progressive refinement network},
  author={Yu, Qihang and Zhang, Jianming and Zhang, He and Wang, Yilin and Lin, Zhe and Xu, Ning and Bai, Yutong and Yuille, Alan},
  booktitle={Proceedings of the IEEE/CVF conference on computer vision and pattern recognition},
  pages={1154--1163},
  year={2021}
}

@inproceedings{he2025lotus,
  title={Lotus: Diffusion-based visual foundation model for high-quality dense prediction},
  author={He, Jing and Li, Haodong and Yin, Wei and Liang, Yixun and Li, Leheng and Zhou, Kaiqiang and Zhang, Hongbo and Liu, Bingbing and Chen, YingCong},
  booktitle={International Conference on Learning Representations},
  volume={2025},
  pages={89454--89467},
  year={2025}
}

@inproceedings{ke2024repurposing,
  title={Repurposing diffusion-based image generators for monocular depth estimation},
  author={Ke, Bingxin and Obukhov, Anton and Huang, Shengyu and Metzger, Nando and Daudt, Rodrigo Caye and Schindler, Konrad},
  booktitle={Proceedings of the IEEE/CVF conference on computer vision and pattern recognition},
  pages={9492--9502},
  year={2024}
}

@inproceedings{fu2024geowizard,
  title={Geowizard: Unleashing the diffusion priors for 3d geometry estimation from a single image},
  author={Fu, Xiao and Yin, Wei and Hu, Mu and Wang, Kaixuan and Ma, Yuexin and Tan, Ping and Shen, Shaojie and Lin, Dahua and Long, Xiaoxiao},
  booktitle={European Conference on Computer Vision},
  pages={241--258},
  year={2024},
  organization={Springer}
}

@inproceedings{rombach2022high,
  title={High-resolution image synthesis with latent diffusion models},
  author={Rombach, Robin and Blattmann, Andreas and Lorenz, Dominik and Esser, Patrick and Ommer, Bj{\"o}rn},
  booktitle={Proceedings of the IEEE/CVF conference on computer vision and pattern recognition},
  pages={10684--10695},
  year={2022}
}

@article{blattmann2023stable,
  title={Stable video diffusion: Scaling latent video diffusion models to large datasets},
  author={Blattmann, Andreas and Dockhorn, Tim and Kulal, Sumith and Mendelevitch, Daniel and Kilian, Maciej and Lorenz, Dominik and Levi, Yam and English, Zion and Voleti, Vikram and Letts, Adam and others},
  journal={arXiv preprint arXiv:2311.15127},
  year={2023}
}

@inproceedings{yang2025cogvideox,
  title={Cogvideox: Text-to-video diffusion models with an expert transformer},
  author={Yang, Zhuoyi and Teng, Jiayan and Zheng, Wendi and Ding, Ming and Huang, Shiyu and Xu, Jiazheng and Yang, Yuanming and Hong, Wenyi and Zhang, Xiaohan and Feng, Guanyu and others},
  booktitle={International Conference on Learning Representations},
  volume={2025},
  pages={83048--83077},
  year={2025}
}

@article{zheng2024open,
  title={Open-sora: Democratizing efficient video production for all},
  author={Zheng, Zangwei and Peng, Xiangyu and Yang, Tianji and Shen, Chenhui and Li, Shenggui and Liu, Hongxin and Zhou, Yukun and Li, Tianyi and You, Yang},
  journal={arXiv preprint arXiv:2412.20404},
  year={2024}
}

@inproceedings{yao2025reconstruction,
  title={Reconstruction vs. generation: Taming optimization dilemma in latent diffusion models},
  author={Yao, Jingfeng and Yang, Bin and Wang, Xinggang},
  booktitle={Proceedings of the Computer Vision and Pattern Recognition Conference},
  pages={15703--15712},
  year={2025}
}

@inproceedings{li2026back,
  title={Back to basics: Let denoising generative models denoise},
  author={Li, Tianhong and He, Kaiming},
  booktitle={Proceedings of the IEEE/CVF Conference on Computer Vision and Pattern Recognition},
  pages={36115--36125},
  year={2026}
}

@inproceedings{lin2023adaptive,
  title={Adaptive human matting for dynamic videos},
  author={Lin, Chung-Ching and Wang, Jiang and Luo, Kun and Lin, Kevin and Li, Linjie and Wang, Lijuan and Liu, Zicheng},
  booktitle={Proceedings of the IEEE/CVF Conference on Computer Vision and Pattern Recognition},
  pages={10229--10238},
  year={2023}
}

@inproceedings{huang2023end,
  title={End-to-end video matting with trimap propagation},
  author={Huang, Wei-Lun and Lee, Ming-Sui},
  booktitle={Proceedings of the IEEE/CVF Conference on Computer Vision and Pattern Recognition},
  pages={14337--14347},
  year={2023}
}

@inproceedings{sun2023ultrahigh,
  title={Ultrahigh resolution image/video matting with spatio-temporal sparsity},
  author={Sun, Yanan and Tang, Chi-Keung and Tai, Yu-Wing},
  booktitle={Proceedings of the IEEE/CVF Conference on Computer Vision and Pattern Recognition},
  pages={14112--14121},
  year={2023}
}

@inproceedings{rhemann2009perceptually,
  title={A perceptually motivated online benchmark for image matting},
  author={Rhemann, Christoph and Rother, Carsten and Wang, Jue and Gelautz, Margrit and Kohli, Pushmeet and Rott, Pamela},
  booktitle={2009 IEEE conference on computer vision and pattern recognition},
  pages={1826--1833},
  year={2009},
  organization={IEEE}
}

@inproceedings{erofeev2015perceptually,
  title={Perceptually Motivated Benchmark for Video Matting.},
  author={Erofeev, Mikhail and Gitman, Yury and Vatolin, Dmitriy S and Fedorov, Alexey and Wang, Jue},
  booktitle={BMVC},
  volume={1},
  pages={2},
  year={2015}
}

@inproceedings{peebles2023scalable,
  title={Scalable diffusion models with transformers},
  author={Peebles, William and Xie, Saining},
  booktitle={Proceedings of the IEEE/CVF international conference on computer vision},
  pages={4195--4205},
  year={2023}
}

@article{yang2026genmask,
  title={GenMask: Adapting DiT for Segmentation via Direct Mask Generation},
  author={Yang, Yuhuan and Zhuang, Xianwei and Cai, Yuxuan and Ma, Chaofan and Bai, Shuai and Yao, Jiangchao and Zhang, Ya and Lin, Junyang and Wang, Yanfeng},
  journal={arXiv preprint arXiv:2603.23906},
  year={2026}
}
}
\end{document}